\documentclass[a4paper]{article}

\usepackage{INTERSPEECH2020}
\usepackage{float}
\usepackage{adjustbox}
\usepackage{threeparttable}
\usepackage{caption} 
\title{Analysis of Visual Features for Continuous Lipreading in Spanish}
\name{David Gimeno-Gómez , Carlos-D. Martínez-Hinarejos}
\address{
  Pattern Recognition and Human Language Technologies Research Center,\\
  Universitat Politècnica de València, Camino de Vera, s/n, 46022, València, Spain
  }
\email{dagigo1@dsic.upv.es, cmartine@dsic.upv.es}

\begin{document}

\maketitle

\begin{abstract}

During a conversation, our brain is responsible for combining information obtained from multiple senses in order to improve our ability to understand the message we are perceiving. Different studies have shown the importance of presenting visual information in these situations. Nevertheless, lipreading is a complex task whose objective is to interpret speech when audio is not available. By dispensing with a sense as crucial as hearing, it will be necessary to be aware of the challenge that this lack presents. In this paper, we propose an analysis of different speech visual features with the intention of identifying which of them is the best approach to capture the nature of lip movements for natural Spanish and, in this way, dealing with the automatic visual speech recognition task. In order to estimate our system, we present an audiovisual corpus compiled from a subset of the RTVE database, which has been used in the Albayzín evaluations. We employ a traditional system based on Hidden Markov Models with Gaussian Mixture Models. Results show that, although the task is difficult, in restricted conditions we obtain recognition results which determine that using eigenlips in combination with deep features is the best visual approach.\\

\end{abstract}
\vspace{-2mm}
\noindent\textbf{Index Terms}: lipreading, machine learning, speech technologies, computer
vision, hidden markov models, deep learning

\vspace{-2mm}
\section{Introduction}

During a conversation, our brain is responsible for combining information obtained from multiple senses in order to improve our ability to understand the message we are perceiving. Different studies have shown the importance of presenting visual information in these situations, as well as its relationship with the sounds produced. Principally, we stand out the studies carried out by McGurk and McDonald \cite{01mcgurk1976hearing}, where they demonstrated that if the mouth expression does not match with the emitted sound, the listener was confused, perceiving a sound different from what it really was. Nevertheless, lipreading is a complex task whose objective is to interpret speech when audio is not available. By dispensing with a sense as crucial as hearing, since this signal presents a greater amount of information regarding speech recognition, it will be necessary to be aware of the challenge that this lack presents. 

Our ideal purpose is to build a system capable of imitating the human ability to interpret continuous speech by reading the lips of the speaker. Due to the absence of acoustic cues, some of the main challenges we have to deal with are visual ambiguities and silence modelling \cite{02fernandez2017optimizing, 03thangthai2018computer}. Therefore, an essential factor is to identify a suitable representation that manages to capture the nature of lip movements and how it affects to the recognition quality. Consequently, the central core of this paper deals with an analysis of speech visual features \cite{04matthews2002extraction, 05parekh2019lip}. 

For this comparison, a fixed decoding system must be chosen. In our case, we employed a traditional approach to define the automatic system, in other words, a system based on Hidden Markov Models combined with Gaussian Mixture Models (GMM-HMM), an approach that has been widely used in Acoustic Speech Recognition (ASR) \cite{06gales2008application}. Although this is not the state-of-the-art for speech-related signal recognition, it is an appropriate option for comparing the different possibilities for feature extraction. Unlike in ASR, when we deal with Visual Speech Recognition (VSR) our basic speech unit is not the phoneme, but the one known as the viseme, which is associated with the representation of the phoneme on the visual domain \cite{07fisher1968confusions}. Unfortunately, there is not direct or one-to-one correspondence between them, which causes visual ambiguities. An interesting work \cite{02fernandez2017optimizing} describes a phoneme-to-viseme mapping for Spanish and concludes its usefulness compared to recognition through phonemes directly. However, we decided to establish the phoneme like basic speech unit in our work as many authors have done \cite{08fernandez2017towards, 09thangthai2015improving, 10potamianos2003recent}.

Apart from that, in order to estimate our system, an audiovisual corpus focused on continuous speech has been built from a subset of the RTVE database \cite{11lleida2018rtve2018}, with Spanish being the language to be interpreted. The RTVE database is a well-known database which has been used in the Albayzin evaluations \cite{12lleida2019albayzin}. Taking into account all these aspects, we can integrate our task both in the field of Speech Technologies and Computer Vision.

In relation to the rest of the paper and its organization, we mention that Section 2 provides the context and historical evolution around the task of VSR or Automatic Lipreading Recognition (ALR). Section 3 presents several details regarding the built audiovisual corpus. Then, Section 4 describes the different visual approaches considered in order to represent the nature of lip movements. Section 5 shows the experimental process carried out in our work, as well as certain insights and comments regarding our results. Finally, conclusions and future lines of research are offered in Section 6.

\vspace{-2mm}
\section{State of the art}

In its origins, automatic speech recognition systems focused only on acoustic information, since this signal is more informative to distinguish phonemes \cite{13fernandez2018survey}. Nowadays, these models are powerful systems capable of understanding spoken language with great quality \cite{14chan2015listen}. However, when the acoustic signal is damaged or corrupted, the performance of these systems decline considerably \cite{13fernandez2018survey, 10potamianos2003recent}. Therefore, many authors have studied how the incorporation of visual cues alongside acoustic information cause a significant improvement over interpretations supplied by the system in these situations \cite{10potamianos2003recent, 15dupont2000audio}. Additionally, several studies related to Silent Speech Interfaces (SSI) \cite{16denby2010silent} were carried out to deal with the possible absence of the acoustic signal in the field of Speech Technologies.

In this way, in the last decades there has been an increase in the interest of decoding speech using exclusively the information from the visual channel. As Fernandez-Lopez and Sukno suggest in their review \cite{13fernandez2018survey}, advances achieved by this type of systems have been conditioned, among other reasons, by the evolution reflected over available audiovisual databases. These databases began by tackling simple tasks from alphabet and digit recognition, such as AVLetters \cite{04matthews2002extraction} and CUAVE \cite{17patterson2002cuave} corpora. More recent datasets provide the necessary support to estimate approaches in charge of interpreting spontaneous speech, for instance, \textsl{Lipreading Sentences in the Wild} (LRS) \cite{18chung2017lip}. Regarding Spanish, we stand out the VLRF corpus \cite{08fernandez2017towards}, despite the fact that it differs from our objectives by recording the scenes under controlled conditions and ensuring that speakers strain theirselves to vocalize adequately and expressively. Lately, the CMU-MOSEAS database \cite{19zadeh2020moseas} has been compiled, among other languages, for Spanish. This is an interesting corpus, as it provides a multimodal point of view, supplying information related with the emotions and subjectivity expressed by the speaker.

At the beginning, these tasks were developed under a traditional paradigm, that is, mainly through the well-known HMMs. Since this is our case, we highlight some publications, such as studies carried out by Thangthai, Cox, and Howell, among others \cite{20howell2016visual, 09thangthai2015improving}, where they employed an HMM per phoneme, evaluating both dependent and context-independent models. On the other hand, in relation to Spanish, we mention again the paper developed by Fernandez-Lopez and Sukno \cite{08fernandez2017towards}, where we emphasize their study regarding recognition at the phoneme level over the VLRF corpus. However, the research has gravitated towards Deep Learning technologies. More concretely, end-to-end architectures, formed by combining Long Short Term Memory (LSTM) \cite{21lstm1999} and Convolutional Neural Networks (CNN) \cite{22krizhevsky2012imagenet}, have been the most widely used topology. In fact, Zisserman \cite{18chung2017lip} reached the state-of-the-art in continuous VSR by employing this approach and achieving an error rate of around 50\% at word level.

\vspace{-2mm}
\section{Audiovisual corpus}

As we mentioned above, we have compiled an audiovisual corpus focused on the task of continuous lipreading recognition, where we could find a large number of speakers in a wide range of scenarios, including variations on intra-personal aspects or light conditions. We compiled it from a subset of the RTVE database \cite{11lleida2018rtve2018} which has been employed in Albayzín evaluations \cite{12lleida2019albayzin}. The RTVE database is made up of different programs broadcasted by \textsl{Radio Televisión Española} but, in our work, we compiled the corpus only from the news program 20H broadcast by the \textsl{Canal 24 horas}. In this program, we have selected scenes where a unique speaker talks from different distances to the camera and in diverse scenarios, either inside a record studio or in outdoor locations. Furthermore, the speaker does not always maintain a frontal plane but can sometimes adopt tilted postures. Other details regarding the compiled dataset are shown in Table 1.

\begin{table}[!htbp]
\vspace{-2mm}
\caption{Details regarding the compiled audiovisual corpus.}
\begin{adjustbox}{width=\columnwidth}
\begin{threeparttable}
\begin{tabular}{@{}c|ccc@{}}
\toprule
\midrule
\textbf{Language} & \multicolumn{3}{c}{Spanish} \\
\textbf{Resolution} & \multicolumn{3}{c}{480$\times$270 pixels, 30 frames/second} \\
\textbf{Speakers} & 57 & \textbf{Males:} 17 & \textbf{Females:} 40 \\
\textbf{Duration} & \multicolumn{3}{c}{$\sim$3 hours} \\
\textbf{Utterances} & \multicolumn{3}{c}{2792} \\
\textbf{Vocabulary} & \multicolumn{3}{c}{2885} \\
\textbf{Running words} & \multicolumn{3}{c}{$\sim$35k} \\
\textbf{Phonemes} & \multicolumn{3}{c}{23} \\
\textbf{Words per utterance} & \textbf{Median:} 10  & \textbf{Max:} 62  & \textbf{Min:} 1  \\
\textbf{Phonemes per utterance} & \textbf{Median:} 46 & \textbf{Max:} 270 & \textbf{Min:} 5\\ \bottomrule
\end{tabular}
\end{threeparttable}
\end{adjustbox}
\label{corpusDetails}
\vspace{-4mm}
\end{table}

Before continuing, it is necessary to mention that the resolution of 480$\times$270 pixels refers to the full record scene. Our region of interest, that is, the speaker's mouth, is of variable dimensions. Therefore, we stablished its size in 32$\times$16 pixels.
\vspace{-2mm}
\section{Speech visual features}

In the literature, a large number of approaches regarding the visual representation of speech have been studied. Many authors \cite{04matthews2002extraction, 23lan2010improving, 24shaikh2010lip} have employed traditionals techniques to extract these features, such as Principal Component Analysis (PCA), Active Appearence Models (AAM), or Optical Flow. Moreover, other authors have delegated the responsibility of extracting visual features on neural networks \cite{25fung2018end, 26palevcek2014extraction, 05parekh2019lip}, and more specifically on Convolutional Autoencoders. 

However, there is no consensus or agreement in the literature on what is the best option for the extraction of visual speech features. Therefore, in our work we studied three types of features that we describe in subsections from 4.1 to 4.3. In all the approaches, the use of the OpenCV library \cite{27opencvlibrary} and the Dlib toolkit \cite{28king2009dlib} allowed us to identify 68 facial landmarks. From some of these landmarks, as left part of Figure 1 reflects, we were able to extract our region of interest.
\vspace{-2mm}
\subsection{Geometric features}

 In this first approach, the study of continuous sign language carried out by Hermann Ney and other authors \cite{29koller2015continuous} is taken as a reference. In this way, we defined, thanks to the location of landmarks described above, a set of 18 high-level features, such as width, height, or area of speaker's mouth. However, when the speaker is more or less close to the camera the same metric (for example, mouth's width) would acquire a value in different magnitudes, even in the case of the same mouth posture or physiognomy. For this reason, we decided to locate a more stable region where each of the measured distances can be properly normalized. This is the region highlighted by a larger blue rectangle on the right side of Figure 1. Thus, the resulting geometric features are normalized using the size of this area.

\begin{figure}[H]
  \vspace{-2mm}
  \centering
  \includegraphics[width=0.42\linewidth]{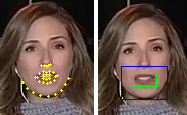}
  \vspace{-1.5mm}
  \caption{Aspects regarding geometric features.}
  \captionsetup[]{width=\linewidth}
  \label{fig:geometricFeats}
  \vspace{-2mm}
\end{figure}

\subsection{Eigenlips}

This concept is influenced by the studies carried out on facial recognition \cite{30delac2005appearance}. Then, as other authors did \cite{03thangthai2018computer}, after computing PCA over our training set we could obtain the eigenlips shown in Figure 2. The first component, as suggests the image, stands out the lip corners, since these are the parts that suffer the greatest deformation throughout a speech. As for the rest of eigenlips, some emphasize lip contours while others highlight zones where we can find teeth or tongue. These are aspects that we can not reach when we work with pure geometric features, but that are of vital importance for visual speech recognition.

\begin{figure}[H]
  \vspace{-2mm}
  \centering
  \includegraphics[width=0.72\linewidth]{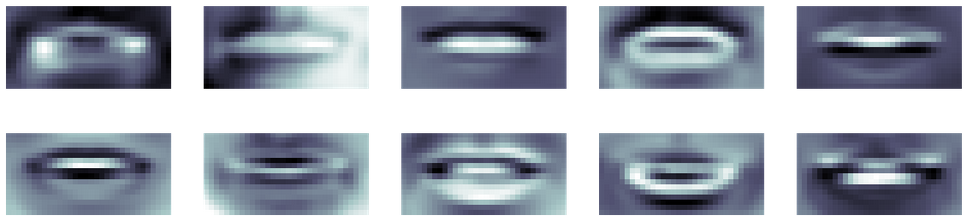}
  \vspace{-1.5mm}
  \caption{Eigenlips obtained after applying PCA.}
  \label{fig:eigenLips}
  \vspace{-2mm}
\end{figure}

Another important issue is that, with the intention of making easier the extraction of these and the features described in Subsection 4.3, we apply a mouth alignment. In other words, as Figure 3 suggests, we rotate those regions of interest where we observe that the speaker's mouth is tilted or inclined.

\begin{figure}[H]
  \vspace{-2mm}
  \centering
  \includegraphics[width=0.4\linewidth]{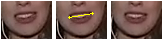}
  \vspace{-1.5mm}
  \caption{Schematic process of mouth alignment.}
  \label{fig:alignMouth}
  \vspace{-2mm}
\end{figure}

\subsection{Deep features}

The last approach, as we mentioned above, is based on Convolutional Autoencoders \cite{31bengio2009learning}, a neural network whose main purpose consists of reconstructing the image received as input from an abstract and compact representation which has been obtained previously from the original image. Once this statistical model is trained, we can dispense with decoder because it is the encoder the component we need to extract our visual features. This scheme is reflected in Figure 4.

\begin{figure}[H]
  \vspace{-2mm}
  \centering
  \includegraphics[width=\linewidth]{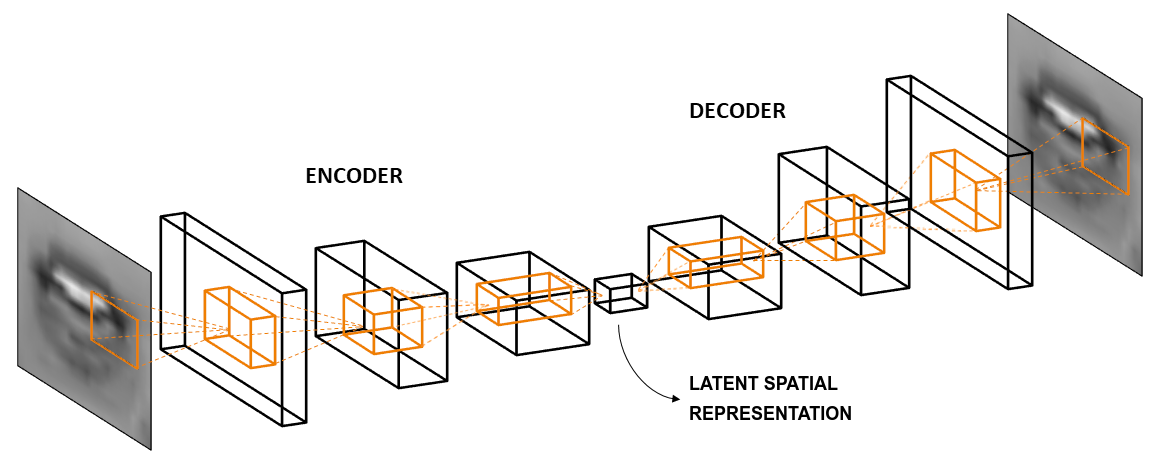}
  \vspace{-1.5mm}
  \caption{Scheme of the proposed Convolutional Autoencoder.}
  \label{fig:autoencoderDeepFeats}
  \vspace{-2mm}
\end{figure}

The encoder architecture is entirely based on that presented in \cite{25fung2018end}, since they employ low-resolution images as is our case. Thus, we were able to obtain high quality reconstruction results, as shown in Figure 5. Finally, we remind that in these features we apply a mouth alignment too.

\begin{figure}[H]
  \vspace{-2mm}
  \centering
  \includegraphics[scale=0.2]{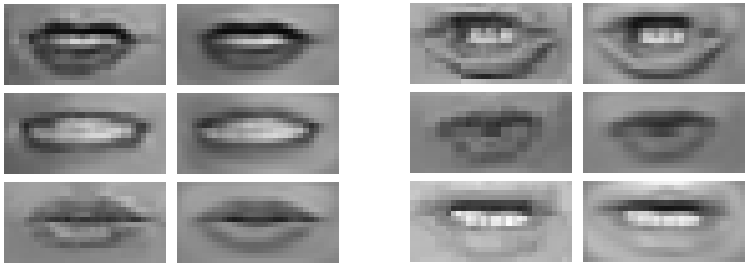}
  \vspace{-1.5mm}
  \caption{Reconstructions examples obtained by the defined Convolutional Autoencoder. For each images block, left column: original image; right column: reconstruction.}
  \label{fig:autoencoderReconstructions}
  \vspace{-2mm}
\end{figure}


\section{Experiments}

\begin{table*}[!htbp]
\vspace{-2mm}
\centering
\caption{Results (WER) for visual speech features with speaker and utterance normalization. \textbf{raw}: refers to the raw features, without adding any type of temporal coefficient. $\Delta$$\Delta$\textsubscript{x}: applies on the features the coefficients \textsl{delta-delta} with a context of \textsl{x} frames. Best result for each set of features and normalization in bold.}
\begin{adjustbox}{width=\textwidth}
\begin{threeparttable}
\begin{tabular}{ccccc|cccc}
\toprule

\multicolumn{1}{c}{} & \multicolumn{4}{c|}{\footnotesize\textbf{{\textsl{z-score} normalization per speaker}}} & \multicolumn{4}{c}{\footnotesize\textbf{{\textsl{z-score} normalization per utterance}}} \\ \midrule
\footnotesize\textbf{Features} & \footnotesize\textbf{raw} & \footnotesize\textbf{$\Delta$$\Delta$\textsubscript{1}} & \footnotesize\textbf{$\Delta$$\Delta$\textsubscript{2}} & \footnotesize\textbf{$\Delta$$\Delta$\textsubscript{3}} &
\footnotesize\textbf{raw} &
\footnotesize\textbf{$\Delta$$\Delta$\textsubscript{1}} & \footnotesize\textbf{$\Delta$$\Delta$\textsubscript{2}} & \footnotesize\textbf{$\Delta$$\Delta$\textsubscript{3}} \\ \midrule
\footnotesize geometricFeats & 51.3$\pm$8.5 & 49.2$\pm$8.2 & \textbf{37.7$\pm$8.2} & 50.5$\pm$7.8 & 46.1$\pm$8.8 & 49.8$\pm$8.6 & \textbf{36.8$\pm$7.1} & 48.7$\pm$8.0 \\
\footnotesize eigenLips & 71.6$\pm$8.6 & 60.6$\pm$8.5 & \textbf{57.1$\pm$8.3} & 65.8$\pm$7.5 & 66.6$\pm$8.8 & \textbf{49.9$\pm$8.1} & 55.6$\pm$8.5 & 53.0$\pm$8.4 \\
\footnotesize deepFeats & 46.6$\pm$8.5 & \textbf{31.7$\pm$7.3} & 35.9$\pm$7.7 & 39.1$\pm$7.8 & 72.4$\pm$7.9 & 58.9$\pm$8.2 & 54.8$\pm$8.9 & \textbf{54.7$\pm$8.2} \\
\footnotesize geometricFeats+eigenLips & 29.6$\pm$7.5 & 30.3$\pm$6.5 & 34.2$\pm$7.2 & \textbf{26.6$\pm$6.0} & \textbf{26.1$\pm$6.2} & 34.6$\pm$7.1 & 31.2$\pm$6.6 & 34.9$\pm$6.4 \\
\footnotesize geometricFeats+deepFeats & 32.9$\pm$7.9 & \textbf{29.8$\pm$7.1} & 34.1$\pm$7.0 & 41.3$\pm$6.7 & \textbf{29.3$\pm$7.7} & 36.5$\pm$7.0 & 33.0$\pm$6.9 & 41.3$\pm$7.6 \\
\footnotesize eigenLips+deepFeats & 45.2$\pm$8.2 & 33.6$\pm$7.9 & \textbf{23.7$\pm$6.4} & 35.4$\pm$7.1 & 38.0$\pm$7.5 & 29.6$\pm$6.8 & \textbf{26.8$\pm$7.2} & 31.0$\pm$7.2 \\
\footnotesize geometricFeats+eigenLips+deepFeats & 30.4$\pm$6.8 & \textbf{27.3$\pm$6.5} & 33.4$\pm$6.5 & 41.5$\pm$6.7 & \textbf{34.6$\pm$7.7} & 36.4$\pm$6.7 & \textbf{34.6$\pm$6.8} & 43.0$\pm$6.8 \\

\bottomrule
\end{tabular}
\end{threeparttable}
\end{adjustbox}
\label{expsPerSpkr}
\vspace{-6mm}
\end{table*}

All our experiments were performed with the Kaldi toolkit \cite{32povey2011kaldi}. This toolkit provides the support to build high performance systems focused on Speech Technologies, from traditional approaches to hybrid models or systems based entirely on Deep Learning. In our case, as we stated in the introduction, we employed a traditional GMM-HMM system. This system is formed by three modules: the Optical Model, in charge of interpreting the pronounced phonemes from visual features sequence; the Lexical Model, responsible for building the words from the phonemes provided by the previous module; and the Language Model, capable of combining the words provided by the Lexical Model with the intention of generating the message interpreted by the system. On the other hand, our corpus is divided into two partitions, allocating to the test set those speakers who did not reach a minimum of seconds. Thus, we get a train set composed of 2672 utterances emitted by 43 different speakers, reaching around 3 hours of data, whereas the test partition comprises 120 samples from 13 speakers, covering 0.13 hours of utterances. Then, due to the limited amount of available data, we had to estimate a context-independent system, also know as monophonic system. 

At the beginning, we carried out several speaker-independent experiments with an Open Language Model, but because of the difficulty of the task, we did not achieve minimally acceptable results (error rates greater than 90\%). Consequently, it was necessary to make experiments in a more constrained scenario in order to obtain conclusions on the use of the different features. Therefore, we decided to relax the task complexity by employing a Closed Language Model. In other words, a Language Model estimated only from the text included in the test partition. In this way, the system has a reduced set of alternatives when it interprets the message, which allows us to focus our experiments on the performance of the different types of features. In fact, as suggested in \cite{08fernandez2017towards}, an acceptable recognition at phonemic level does not necessarily imply a good quality performance when word level decoding message is carried out.  

Our first experiment focused on studying HMM's topology, one of the most relevant factors in relation to temporal alignment of visual data. The classic topology in ASR (3 states left-to-right with self-loops) provided us a poor recognition rate. Then, employing raw geometric features, we tested several topologies. After these experiments, we observed that if we reduce the number of states and we add transitions of each of these states to the final state, system performance obtains, in general, a considerable improvement. According to these results, we believe that this behaviour may be caused by the limited frequency of information (30 frames/second) that presents the visual channel with respect to the standard representation (Mel Frequency Cepstral Coefficients, MFCC) of acoustic data (100 frames/second) \cite{06gales2008application}. Consequently, in the rest of our work we employed the topology shown in Figure 6: 2 states left-to-right with self-loops and skip transitions.

\begin{figure}[H]
  \vspace{-2mm}
  \centering
  \includegraphics[scale=0.15]{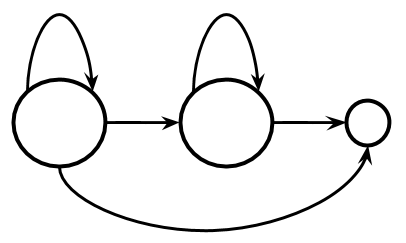}
  \vspace{-1.5mm}
  \caption{HMM's topology employed in our experiments.}
  \label{fig:hmmTopologyD}
  \vspace{-2mm}
\end{figure}

Once this aspect is specified, we can address the analysis of the visual speech features with the intention of determining which of them, either isolated or combined with each other, is the best representation to capture the nature of lip movements. In addition, we study how the incorporation of temporal or delta-delta coefficients, used in ASR \cite{06gales2008application}, affects to the recognition quality. More precisely, we study different contexts, both a greater and a lower scope. On the other hand, as Lan presented in \cite{23lan2010improving}, we have observed how applying a z-score normalization causes a considerable recognition quality improvement. Then, in our analysis we studied two types of normalization on all the features presented in Section 4:   

\begin{itemize}
    \item \underline{Normalization per speaker:} all the utterances of a speaker are taken in the normalization; this is aimed at mitigating the differences that may exist in the aspect of the speaker in his/her different utterances (light conditions, facial hair, lipstick colour, temporal scars or marks in the mouth, ...). 
    
    \item \underline{Normalization per utterance:} the normalization is for each single utterance; this is done with the intention of reducing the differences among the different utterances from different speakers (i.e., to obtain speaker independency) and conditions (e.g., focus variability). 
\end{itemize}

All the results are evaluated by the well-known Word Error Rate (WER) with 95\% confidence intervals obtained by the boostrap method described in \cite{33bisani2004bootstrap}.

Results are presented in Table 2. First, we stand out that the incorporation of delta-delta coefficients cause, in general, an improvement on system performance. Of course, depending on the type of features or normalization, it is convenient to use one context or other. 

On the other hand, if we focus only on those experiments that study the visual features individually, we conclude that deep features are the best representation, in isolated way, to address VSR, as long as we normalize per speaker. In contrast, when we apply a normalization per utterance, we notice how the quality of these features suffer a drastic deterioration, while the rest of visual approaches improve their results. This may mean that deep features, as they are highly dependent on the pixel values, are more affected by changes in light conditions or certain intra-personal aspects. In this way, if these features are processed by a normalization per speaker, they are more benefited in order to interpret speech visually. In contrast, geometric features are more dependant on the location of the specific pixels, and thus utterance normalization fits better for these features. Eigenlips depend on both specific pixels and their values, which makes it to be the worst option when normalizing by speaker and to not improve geometric features when normalizing by utterance.

Regarding the experiments that explore the combinations of visual features, we confirm that, as a general rule, feature combination produces a decrease in error rates. Therefore, we can deduce that the studied features complement each other and manage to provide a more robust representation. Nevertheless, this is not always the case; in certain occasions these results are overlapped with the best error rates obtained in experiments where features were employed individually. On the other hand, the eigenlips and deep features combination, if we normalize per speaker, is established as the best approach to address the automatic lipreading, reaching around 23.7\% WER, although differences are not significant with respect to only using deep features. Seemingly, thanks to appearance aspects contained in eigenlips and the great potential that deep features demonstrated regarding mouth physiognomy reconstruction, a high quality representation has been achieved. However, it is worth noting that if we do not incorporate delta-delta coefficients, we can verify how the geometric features and eigenlips combination improves the performance of the previously mentioned approach.   

Finally, we stand out that the combination of all the features analyzed in our work forms a representation with a large dimension. This fact can cause difficulties when modelling data statistically. Consequently, unlike most cases in which isolated features are used, introducing temporal coefficients does not always imply better results.
\vspace{-2mm}
\section{Conclusions}

In this paper, an extensive study has been carried out regarding visual speech features in order to address an automatic lipreading task for natural Spanish. Therefore, in addition to compiling a preliminary audiovisual corpus, approaches based on both traditional techniques and Deep Learning architectures have been addressed to represent the nature of lip movements. After our experiments, we conclude that the combination of eigenlips and deep features, as long as we apply certain aspects such as normalization and temporal coefficients, provide the best approach to interpret speech visually. On the other hand, aspects in relation to temporal alignment of visual data have been studied. More concretely, according to the results obtained in this work, the HMM's topology has been modified regarding standards in acoustic speech recognition. 

However, visual speech recognition remains an open problem. Lipreading is a complex task where researchers have to face with visual ambiguities and other issues such as silence modelling. In addition, there is not a consensus or agreement regarding a suitable visual speech representation. For these and others reasons, further research is necessary. Authors such as Fernandez-Lopez and Sukno \cite{13fernandez2018survey} have suggested that future lines of research should be developed around temporal alignments and context modelling. On the other hand, we consider to shift our study towards a pure Deep Learning approach. More precisely, we aim at an end-to-end architecture whose parameters, including those in charge of extracting visual features, are estimated according to the mistakes identified in the message decoding stage. In other words, we believe that this direct learning could be more useful than addressing the task with a traditional approach, where each module is independent of the other. In order to achieve this objective, we must increase the number of seconds which forms the compiled audiovisual corpus. In this way, we expect to get a large amount of data which represents the nature of natural speech and be able to estimate our statistical models appropriately. Finally, it would be interesting to study whether a suitable viseme-phoneme correspondence for Spanish can lead to advances in the matter.
\vspace{-1.5mm}
\section{Acknowledgements}

This work was partially supported by Generalitat Valenciana under project DeepPattern (PROMETEO/2019/121) and by Ministerio de Ciencia under project MIRANDA-DocTIUM (RTI2018-095645-B-C22).
\vspace{-1.5mm}




\bibliographystyle{IEEEtran}

\bibliography{main}

\end{document}